\documentclass[letterpaper]{article} 
\usepackage{aaai24}  
\usepackage{times}  
\usepackage{helvet}  
\usepackage{courier}  
\usepackage[hyphens]{url}  
\usepackage{graphicx} 
\usepackage{natbib}
\usepackage{caption}
\usepackage{algorithm}
\usepackage{algorithmicx}
\usepackage[noend]{algpseudocode}
\usepackage{mathtools}  
\usepackage{booktabs}

\usepackage{newfloat}
\usepackage{listings}

\usepackage{amsfonts}

\DeclareCaptionStyle{ruled}{labelfont=normalfont,labelsep=colon,strut=off} 
\floatstyle{ruled}
\newfloat{listing}{tb}{lst}{}
\floatname{listing}{Listing}

\title{RamseyRL: A Framework for Intelligent Ramsey Number Counterexample Searching}

\author{
    Steve Vott\textsuperscript{\rm 1}\equalcontrib,
    Adam Lehavi\textsuperscript{\rm 1}\equalcontrib,
}
\affiliations{
    \textsuperscript{\rm 1}University of Southern California\\
    svott@usc.edu, alehavi@usc.edu
}

\begin{document}

\maketitle

\begin{abstract}

The Ramsey number is the minimum number of nodes, $n = R(s, t)$, such that all undirected simple graphs of order $n$, contain a clique of order $s$, or an independent set of order $t$. This paper explores the application of a best first search algorithm and reinforcement learning (RL) techniques to find counterexamples to specific Ramsey numbers. We incrementally improve over prior search methods such as random search by introducing a graph vectorization and deep neural network (DNN)-based heuristic, which gauge the likelihood of a graph being a counterexample. The paper also proposes algorithmic optimizations to confine a polynomial search runtime. This paper does not aim to present new counterexamples but rather introduces and evaluates a framework supporting Ramsey counterexample exploration using other heuristics. Code and methods are made available through a PyPl package and GitHub repository.
\end{abstract}

\section{Introduction}

Ramsey theory is a branch of the mathematical field of combinatorics that focuses on the appearance of order in a substructure given a structure of a known size. This paper focuses on 2-color Ramsey numbers, using the definition provided in the abstract. The list of all known Ramsey numbers is regularly updated, and continuously improved upon \cite{radziszowski2021small}, where some Ramsey numbers are not known explicitly but have known bounds.

Ramsey theory has many interesting applications in the fields of math and computer science, illuminating the emergence of order within large structures. Some applications include those concerning parallel search and computation, complexity classes, concrete complexity, logic, and computational geometry, as compiled by William Gasarch \cite{gasarch2021applications}.

For unknown Ramsey numbers, the main approaches towards finding the Ramsey number typically fall into three categories. The first is direct proofs, that aim to show that a given Ramsey exists through deterministic methods \cite{ronald1990ramsey}. The second is statistical bounding, where intelligent estimation and bounds are used to limit the Ramsey number based on the number of edges or certain other measures related to the graph \cite{nesetril2012mathematics}. The last is finding counterexamples. In this paper we use the terms ``counters" and ``counterexamples" interchangeably. A counterexample $R(s, t, n)$ is one that proves that $R(s, t) > n$, by giving a graph of size $n$ that does not satisfy the conditions of $R(s, t)$. While this method can only reduce the lower bound, it has been shown to have success in the past \cite{chen1993graphs, su1999new, kim1995ramsey}. Searching for all counters entails finding multiple graphs, where all graphs listed must be counterexamples not isomorphic to one another.

One of the ideas behind finding counters is using an intelligent heuristic. This means that instead of searching over all graphs, an intelligent agent is used to search a subset based on a scoring metric. In this way, an agent interacts with an environment, and gets a score returned, similar to reinforcement learning \cite{sutton2018reinforcement}. Reinforcement learning has been used in graph theory in the past and shown positive results over large search spaces when chosen intelligently \cite{wagner2023finding}.

In the field of graph theory, graph invariants define numerical abstractions or vectorized representations of graphs such that two isomorphic graphs always have the same invariant value. It is not inherently true that an equal invariant between two graphs indicates isomorphism, but with an intelligent invariant, the odds of such a claim being true are high \cite{grohe2020graph}. Examples of invariants include the degree sequence of a graph, and invariants can serve as a way to take the problem of isomorphism, which is yet to be solved in polynomial time, and abstract it to a polynomial-time estimation \cite{babai2016graph, grohe2020graph}. This can be used to intelligently search complex graph problems.

\section{Literature Review}
\subsection{Finding Known Bounds}

An updated list of all known Ramsey number bounds is tracked by the Electronic Journal of Combinatorics, and updated regularly \cite{radziszowski2021small}. This list covers various Ramsey numbers, including the most intuitive two-color complete graph Ramsey numbers that we seek to improve upon. This list also links to all past methods for most current bounds, as well as papers referring to applications and datasets. One of the main places where one can find the largest active counterexamples is \cite{ramsey_graphs}, which is a compilation of data sources from many different sources. This compilation include notes on bounds with found counterexample sets that may have additional counterexamples or can be improved to higher \textit{n} values. $R(5,5,42)$ and $R(4,6,35)$ are both examples of known bounds that may have undiscovered counterexamples \cite{ramsey_graphs}. These graphs can also be taken as starting points and used in the search of improved bounds, which we do for $R(4,6,36)$.

In the paper in which Geoffrey Exoo discovers $R(4, 6, 35)$ counterexamples, he introduces 3 methods for computational searches \cite{exoo2012ramsey}. Methods A and C start with a random and symmetrically colored graph, respectively, and then modify the graph edge by edge. This process continues until all bad subgraphs are eliminated, and then \textit{n} is incremented. This approach starts with small enough \textit{n} values to allow for easy search of these symmetries and to quickly eliminate bad subgraphs, and from there increases in \textit{n} until a nontrivial bound is found. The approach for choosing which edge is simulated annealing or similar probabilistic pseudo-random searches. Method B begins with a graph of the desired \textit{n}, and from there uses an intelligent heuristic to choose which edge to change. Instead of just minimizing bad subgraph counts, when looking at the vectorization of all 11 subgraphs of size 4 for a given graph, it seeks to maximize monochromatic 4paths. In this study, the vectorizations were produced with partial human oversight, limiting the framework's direct scalability to larger graphs. Moreover, the sole search methodology employed was simulated annealing; other potential strategies, such as examining graphs an edge apart, were not explored.

\subsection{Reinforcement Learning and Combinatoric Problems}

In ``Constructions in combinatorics via neural networks," Wagner \cite{wagner2021constructions} uses reinforcement learning and the deep cross-entropy method to form graph constructions for a variety of combinatorics problems. Specifically, he translates a combinatorial problem into a problem about generating a word of certain length from a finite alphabet, and uses a neural network to guide decision-making. In each iteration of his algorithm, he inputs sequence $w$ into a neural network receiving a probability distribution on the next letter. Letter $x$ is sampled from this distribution and appended to sequence $w$. This algorithm iterates until sequence $w$ is complete at which point the neural network is trained to minimize the cross-entropy loss over sequences that maximize a desired score. In this paper, the main future direction pointed out is to use a different reinforcement algorithm than deep cross-entropy for an open problem in combinatorics. Additionally, the approach to solving avoids complex multi-step problem formulations, as those are not the ideal candidates for deep cross-entropy \cite{lapan2018deep}.

Markov Decision Processes (MDPs) offer a powerful framework for modeling decision-making problems where an agent interacts with an environment to achieve a specific objective \cite{puterman1994markov}. An MDP is typically characterized by a tuple consisting of states, actions, transition probabilities, and rewards. The agent observes a state, takes an action, receives a reward, and transitions to a new state. The goal is to find a policy, a mapping from states to actions, that maximizes some measure of long-term reward, typically the expected cumulative reward \cite{bertsekas1995dynamic}. In recent years, the applicability of MDPs has expanded beyond traditional realms, finding relevance in various complex domains, including graph search problems. In the context of counterexample search for Ramsey numbers, the formulation as a deterministic MDP provides a structured approach to efficiently navigate the space of potential graph configurations.

Graph Neural Networks (GNNs) have emerged as a powerful paradigm for learning on graph-structured data, capturing attention due to their ability to model intricate systems represented as graphs \cite{bronstein2017geometric}. These methods are fundamentally based on the principle of message passing, a process by which nodes collect and aggregate information from their proximate neighborhood, updating their own features \cite{gilmer2017neural}. The capacity to gather local context through iterative message passing layers is reminiscent of the extraction of graph invariants, attributes that remain unchanged under graph isomorphisms \cite{xu2018powerful}. Such invariants, or higher-level features, serve to encapsulate the graph's complex structure into succinct, fixed-size representations. A crucial aspect of these invariants pertains to graph vectorizations, where nodes, edges, and their associated features are projected into a lower-dimensional space, preserving essential properties of the original graph \cite{perozzi2014deepwalk, tang2015line}. In practice, these vectorizations are used as input features for neural networks, facilitating various tasks including node classification, link prediction, or guiding search strategies within the graph. The utility of GNNs is encapsulated in their ability to generate these concise yet expressive representations of graph structures, enabling the application of robust machine learning (ML) techniques on graph data.

\section{Methodology}
The Ramsey number counterexample search problem can be naturally formulated as a deterministic Markov Decision Process (MDP) with the definitions below.

\begin{itemize}
    \item \textbf{State Space:} A state $S$ is defined as a given graph $G$, reward $r$, and all prior explored graphs $G_{prior}$. Graphs are stored and compared by their vector representation.
    \item \textbf{Actions:} An action is dictated by adding or removing an edge $e$, such that graph $G$ becomes $G'$. Reward $r'$ is also then updated to either be $\alpha r$ if $G'$ is not a counterexample, or $1$ if it is. $G$ is added to $G_{prior}$. Given the use of $G_{prior}$, executing an action in a given state will always lead to a specific new state.
    \item \textbf{Reward Function:} The reward function estimates the likelihood of a graph (represented by its vectorization) being a counterexample. This likelihood is assessed using our neural network heuristic, which has been trained to approximate how close a graph is to being a counterexample. The neural network heuristic is continually updated over batches of iterations. An additional parameter \( \alpha \), which was set to 0 for our experiments, allows for the possibility to score a graph that is not a counterexample as \( \alpha^k \), if the last counterexample was found \( k \) iterations away. However, a nonzero \( \alpha \) was observed to hinder escaping from local maxima, leading to its nullification in our tests.
    \item \textbf{Policy Function:} The objective is to determine an optimal policy that dictates the best action to take in every state to maximize the cumulative reward, which in this case is finding a counterexample. This policy is deterministic, and its approximation is achieved using the neural network heuristic. The Best First Search Algorithm, detailed in algorithm \ref{algorithm_1}, serves as the strategy to explore the state space and refine this policy. The approach herein differs from Wagner's \cite{wagner2021constructions} in its usage of an immediate reward system and vectorization over graph invariants. It is important to note that we deviated from traditional methods of solving an MDP like Q-learning and Dynamic Programming (DP). These methods optimally solve finite MDPs. While our MDP is finite, the exponential size of our state space makes finding an optimal policy computationally intractable, leading us to use a best first search approach.
\end{itemize}

\algnewcommand\algorithmicinput{\textbf{Input:}}
\algnewcommand\Input{\item[\algorithmicinput]}
\DeclarePairedDelimiter\parens()

\begin{algorithm}[H]
    \caption{Best First Search}\label{algorithm_1}
    \begin{algorithmic}
        \Input A model $H$, a graph $G$, a set of edges $E$
        \State training\_data $\gets []$
        \State past $\gets []$
        \While {$G$ exists}
            \State new\_graphs $\gets []$
            \For {$e \in E$}
                \State $G' \gets $Change\_edge$(G,e)$
                \If {$G'$ is a counterexample}
                    \State save $G'$
                \EndIf
                \State $V \gets $vectorize$(G')$
                \If {$V \not \in$ past}
                    \State new\_graphs.append$(V, G')$
                    \State training\_data.append$(V)$
                    \State past.append$(V)$
                \EndIf
            \EndFor
            \State $V, G = \max_{\parens*{V_i,G_i} \in \text{new\_graphs}} H(V_i)$
            \If {$|$training\_data$| =$ batch\_size}
                \State train H on training\_data
                \State training\_data $\gets []$
            \EndIf
        \EndWhile
    \end{algorithmic}
\end{algorithm}

\subsection{Starting Graph}
While Exoo \cite{exoo2012ramsey} attacks lower bounds with 3 methodologies, we parameterize a general approach. In exploring a particular $R(s, t, n)$ counterexample, the initialized graph can vary between an empty graph, a random graph, a prior $R(s, t, n')$ counterexample where $n' < n$, or a current $R(s, t, n)$ counterexample. If using a prior counterexample, a new node is added with edges to all other nodes randomly connected. Only the edges between this new node and other nodes are editable in this case, as the rest of the graph must inherently form a counterexample already. This reduces the search space from $2^{\binom{n}{2}}$ to $2^n$ total graphs when using a prior graph. 

We measured our deep neural network against both a random heuristic and a best-first search implementation of Exoo's proposed 4Path heuristic \cite{exoo2012ramsey}. The random heuristic and random starting graphs both underperform, as expected, and so are not shown in results. Throughout this paper, ``steps" and ``iterations" are used interchangeably, with both terms denoting the execution of a single action as per algorithm 1 to transition the graph and vectorization to a new state. Our reward model is trained on all graphs with new vectorizations per batch of steps, to avoid overfitting to seen examples from old batches. This lowers training time and leads the model to forget less relevant graphs. This is done with the motivation of escaping local maxima and exploring more graphs.

\subsection{Vectorization}

Inspired by Exoo's 4-node subgraph counts of $R(4, 6, 35)$ counterexamples, we designed our vectorization to hold the counts of all 4-node subgraph structures, n, s, t, and a Boolean on the graph being a valid $R(s, t, n)$ counterexample \cite{exoo2012ramsey}. When training a scaled deep neural network (SDNN), it means that the counts of all these subgraph structures are normalized such that they sum to 1. Much like Exoo, we believe these 4-node subgraph counts reveal the structure to a particular counterexample. While Exoo had to manually inspect and set his own goal vectorizations to trend towards, by not having that aspect of human supervision, our model can aim to learn the most important values. 

We consider the counts of 4-node subgraph structures as graph invariants. Therefore, this vectorization holding 11 different graph invariants is in itself, a strong graph invariant. We use this property to turn our isomorphism checking into a vectorization checking whereby we guarantee exploring new graphs by only exploring new vectorizations.

\subsection{Vectorization Runtime Analysis}
The size of our search space is essentially the number of simple graphs over $n$ nodes: $2^{\binom{n}{2}} = 2^{n(n-1)/2}$. When starting from a prior counterexample and bounding the explorable edges, we get a search space of $2^n$ possible graphs. Although the search space is exponential, our heuristic can converge towards valid counterexamples in a polynomial number of steps in theory, and often does so in practice. 

We bounded our vectorization runtime of 4-node subgraph counts from $\binom{n}{4}$ to $\binom{n}{2}$ when starting from prior counterexamples by taking advantage of the fact that each step only updates a single edge. We detail this with algorithm \ref{algorithm_2}.

\begin{algorithm}[H]
    \caption{Efficient Vectorization}\label{algorithm_2}
    \begin{algorithmic}
        \Input A graph $G$, edge $e(u,v)$, prior Vectorization $V$
        \State Old count O\_C $\gets$ vec\_w\_e$(G, u, v)$
        \State $G' \gets $Change\_edge$(G,e)$
        \State New count N\_C $\gets$ vec\_w\_e$(G, u, v)$
        \State New Vectorization N\_V$\gets$ \{\}
        \For {struct $\in$ 4-node\_Structs}
            \State N\_V[struct] $\gets$ V[\text{{struct}}] $\!+\!$ (\text{N\_C[{struct}]} $\!-\!$ \text{O\_C[{struct}]})

        \EndFor
        \State \textbf{return} N\_V
    \end{algorithmic}
\end{algorithm}

    Our vectorization is implemented as a dictionary of \{4-node subgraph structure: count\} pairs: e.g. \{K\_3: 10\} means our graph has 10 4-node subgraphs that hold a K\_3 and lonely node, where K\_3 is a complete subgraph of order 3. We obtain a vectorization, O\_C, for all 4-node subgraphs with nodes in our changed edge. We then change the edge and obtain a vectorization N\_C with the same technique. Vec\_w\_e iterates over all pairs of nodes $\langle w,x \rangle$ in the set of nodes $V \backslash \{u, v\}$, then counts subgraph structures for $\{w,x,u,v\}$ and logs these codes in the vectorization. To figure out our N\_V after changing edge $e(u,v)$ we know that only the 4-node subgraphs that contain $u,v$ can change subgraph counts, as all other 4-node subgraphs without $e(u,v)$ will have the same structure and thus the same subgraph count. Therefore, we can obtain these counts for both the original graph and the graph with changed edge $e(u,v)$, and then assign our N\_V counts to be the old vectorization plus the difference.

We applied the same idea of using our vectorization and current actionable edge to optimize our counterexample check. In classifying a graph as an $R(s, t, n)$ counterexample, we have to iterate over all possible $s$-cliques and independent sets of size $t$ - denoted as $I_t$. For Ramsey number $R(s,t,n)$ this incurs a runtime of $O(n^{\max(s,t)})$. However, because we maintain our graph vectorization throughout, we already have information on the number of $s$-cliques and $I_t$ for $s, t \leq 4$. Such information reduces the runtime to a simple constant comparison on the existence of any $s$-cliques/$I_t$ in our vectorization. In fact, this runtime optimization holds for any $s, t$ when our vectorization counts all $k$-node subgraph structures where $k \geq \max(s,t)$.

In a general sense, the main downside of an approach like this is the lack of subgraph information when $s$ or $t$ exceed $k$. However, algorithm 1's run information is logged through a software called Neptune.ai. This gives insight into how close a given run is to finding a counterexample, beyond just a vectorization. If a given step takes more time, it can be assumed that more graphs reach that $O(n^{\max(s,t)})$ runtime in counterexample classification at which point the number of undesirable subgraphs of order $\max(s,t)$ has an inverse relationship with our runtime. The sooner we find an undesirable subgraph, the sooner we exclude a graph from counterexample consideration. So, although we lack subgraph information when $s$ or $t$ exceed our $k$-node vectorization, we can loosely extract such information using a given step's runtime. Where past approaches explicitly counted all bad subgraphs of order $s,t$ our approach uses logging analysis.

\subsection{Neural Network}

Our parameter and logging choices were also motivated by Exoo's work that required personal inspection. As a simple heuristic succeeded, we aimed to take our vectorization of size 14 (11 scaled subgraph counts, n, s, t) and pass it through a simple DNN. The one we used had  hidden layers of size 36, 12, and 1. We additionally allowed our model to pretrain, but only provided a sparse amount of training data from significantly lower \textit{n} values, so that the model has a motivation to predict nonzero values. We introduced 11,505 labelled graphs and their respective vectorizations to our neural network. The training data used were counterexamples for Ramsey numbers $R(3, t)$ where $t \in \{4,5,6,7,9\}$ along with all isomorphic graphs with 5 or 6 nodes. Counterexamples are also not loaded into a list for comparison when solutions are found. The motivation behind all of this is to prove the validity of the framework with as lightweight and unsupervised of a technique as possible. 

\subsection{Complete Runtime Analysis}
The lowest bound for the number of steps to finding any $R(s, t, n)$ from a graph of size \textit{n} is at most changing each edge once, so $\binom{n}{2} = \Omega (n^2)$. In our implementation, assuming a perfect agent and heuristic, this is the ideal number of steps. Assuming a completely flawed agent, the upper bound is the entirety of the state space, which is $2^{\binom{n}{2}}$.

Examining the runtime of each step, there are $\binom{n}{2}$ graphs one edge away to check. If a graph explicitly is not a counterexample, either by its vectorization or by an edge estimate, it takes $\theta (n^4)$ to consider that graph. If a graph instead can be a counterexample on the basis of having no explicit bad subgraphs in the vectorization, it takes $O(n^{\max(s,t)})$ and $\Omega (n^4)$ time.

So, for a given step, the runtime is $\Omega (n^6)$ and $O(n^{\max(s,t) + 2})$. This means that, in the case of $R(4, 6)$, we expect the ideal approach to take $O(n^{10})$ time, with $O(n^8)$ and $\Omega (n^6)$ time per step. While this is by no means a low degree polynomial, it is of polynomial degree. Additional optimizations can be added to further decrease the worst-case runtime or to balance the amortization better.

While unmeasured, we applied the theorems and computational findings of Goedgebeur and Radziszowski \cite{goedgebeur2012new} to exclude certain graphs from being considered as counterexamples based on their edge counts. Specifically we used their  $e(3,k,n)$ table, where $e(3,k,n)$ is the minimum number of edges required for a graph to be a $R(3,k,n)$ counterexample, and two bounds $e(3,k+1,n) >= (40n-91k)/6$ and $e(3,k+1,n) >= 6n-13k$ . Using their contribution helps reduce the counterexample classification into a constant comparison for some cases.

\section{Results}
All code was run on Google Colab notebooks with the default Intel Xeon CPU with 2 vCPUs (virtual CPUs) and 13GB of RAM. Since our runtime is CPU bounded in our combinatorial subgraph checking we did not use any GPUs for neural network speedup.

\subsection{Time per Iteration vs. Counters Found}
\begin{figure}[ht]
\centering
\includegraphics[width=0.9\columnwidth]{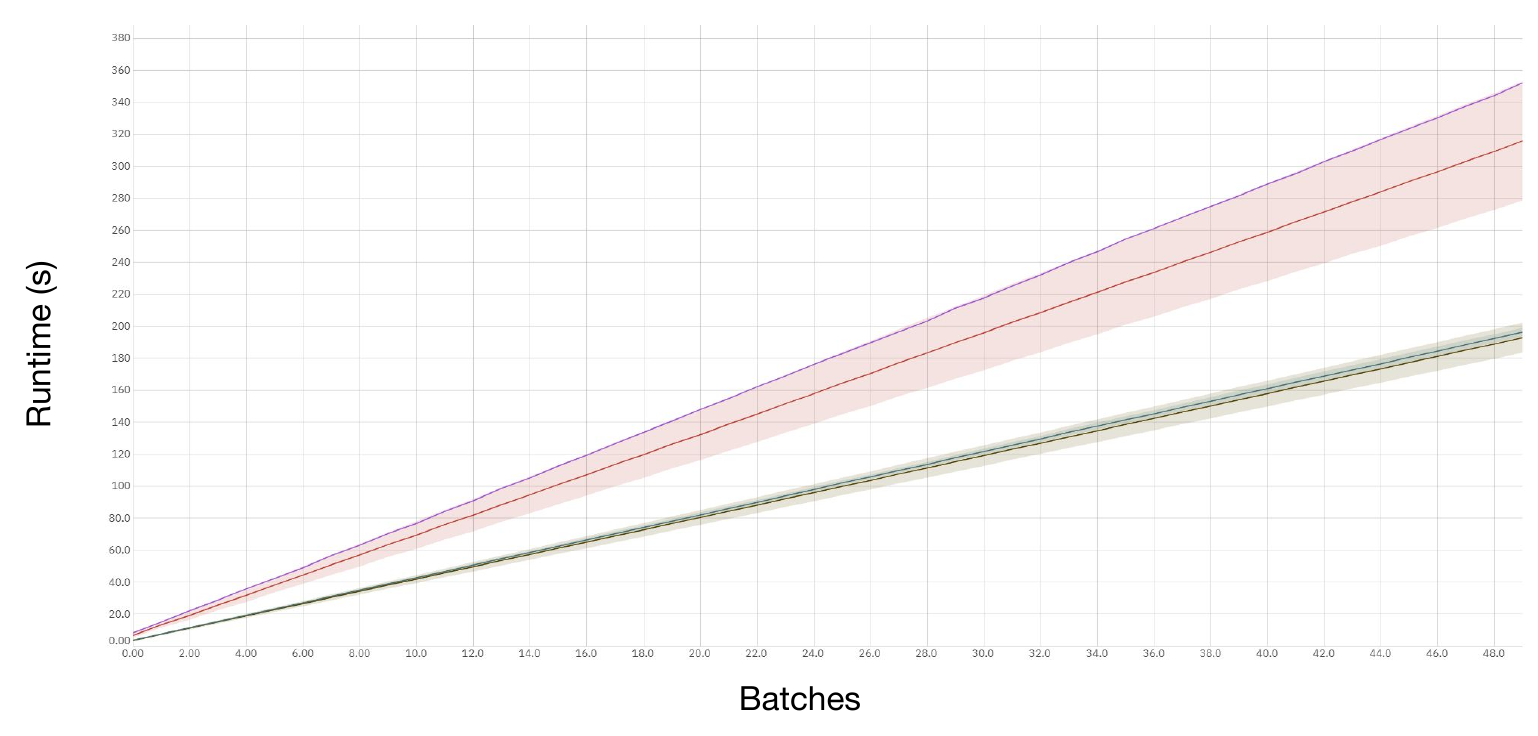}
\caption{Runtime vs iterations for $R(3, 5, 11)$ when grouped by number of counters and heuristic. The order of the lines from slowest to fastest is Scaled DNN with counters found, 4PATH with counters found, Scaled DNN without counters found, and 4PATH without counters found}
\label{running_time_v_counters}
\end{figure}

The first result pertains to the time of a run and iterations in relation to the number of counters found. Figure \ref{running_time_v_counters} shows an example for $R(3, 5, 11)$, with shaded regions indicating the range of all runs grouped. When grouped by heuristic, we noticed a slower runtime for runs with counters found over those with none found. This is attributed to our efficient vectorization technique where our runtime increases as we approach valid counterexamples. I.e. valid $R(s, t, n)$ counterexamples will not have $s$-cliques, making our counterexample checking go from a constant comparison to a runtime polynomial in $t$ for Ramsey numbers with $s \leq 4, t \geq s$.

\subsection{All Runs}

Of the experiments run in table \ref{tab:all_runs}, all show 4PATH excelling as a heuristic when compared to SCALED\_DNN. However, variance for the 4PATH is notably higher than it is for SCALED\_DNN, indicating outliers pushing the average calculated by mean above that of SCALED\_DNN. As an example, for $R(4, 5, 24)$, variance is notably higher for 4PATH. Upon further investigation of runs with the same starting\_graph\_index for $R(4, 5, 24)$, 50\% had more counters for 4PATH, but 17\% had more for SCALED\_DNN and 33\% were equal.
\begin{table}[ht]
    \begin{tabular}{lllllll}
    \hline
    N & S & T & Heuristic & Counters & Variance & Runs \\ \hline
    11 & 3 & 5 & 4PATH & 0.04 & 0.04 & 105 \\
    11 & 3 & 5 & SDNN & 0.03 & 0.03 & 40 \\
    14 & 4 & 4 & 4PATH & 198.73 & 14388 & 116 \\
    14 & 4 & 4 & SDNN & 19.13 & 610.01 & 39 \\
    15 & 4 & 4 & 4PATH & 15.90 & 66.05 & 103 \\
    15 & 4 & 4 & SDNN & 7.00 & 3.07 & 44 \\
    21 & 3 & 7 & 4PATH & 0 & 0 & 33 \\
    21 & 3 & 7 & SDNN & 0 & 0 & 17 \\
    24 & 4 & 5 & 4PATH & 29.16 & 844.89 & 51 \\
    24 & 4 & 5 & SDNN & 14.96 & 25.86 & 23 \\
    36 & 4 & 6 & SDNN & 0 & 0 & 14 \\
    $\forall$ &  &  &  &  &  & 587 \\ \hline
    \end{tabular}
    \caption{A table of run results for 587 runs done for 1000 iterations over both 4PATH and SCALED\_DNN (abbreviated as SDNN), showing all runs fully completed without error. Counters is the average \# of counters, variance is the counterexample variance.}
    \label{tab:all_runs}
\end{table}

\subsection{Runtime Results}

\begin{figure}[ht]
\centering
\includegraphics[width=0.9\columnwidth]{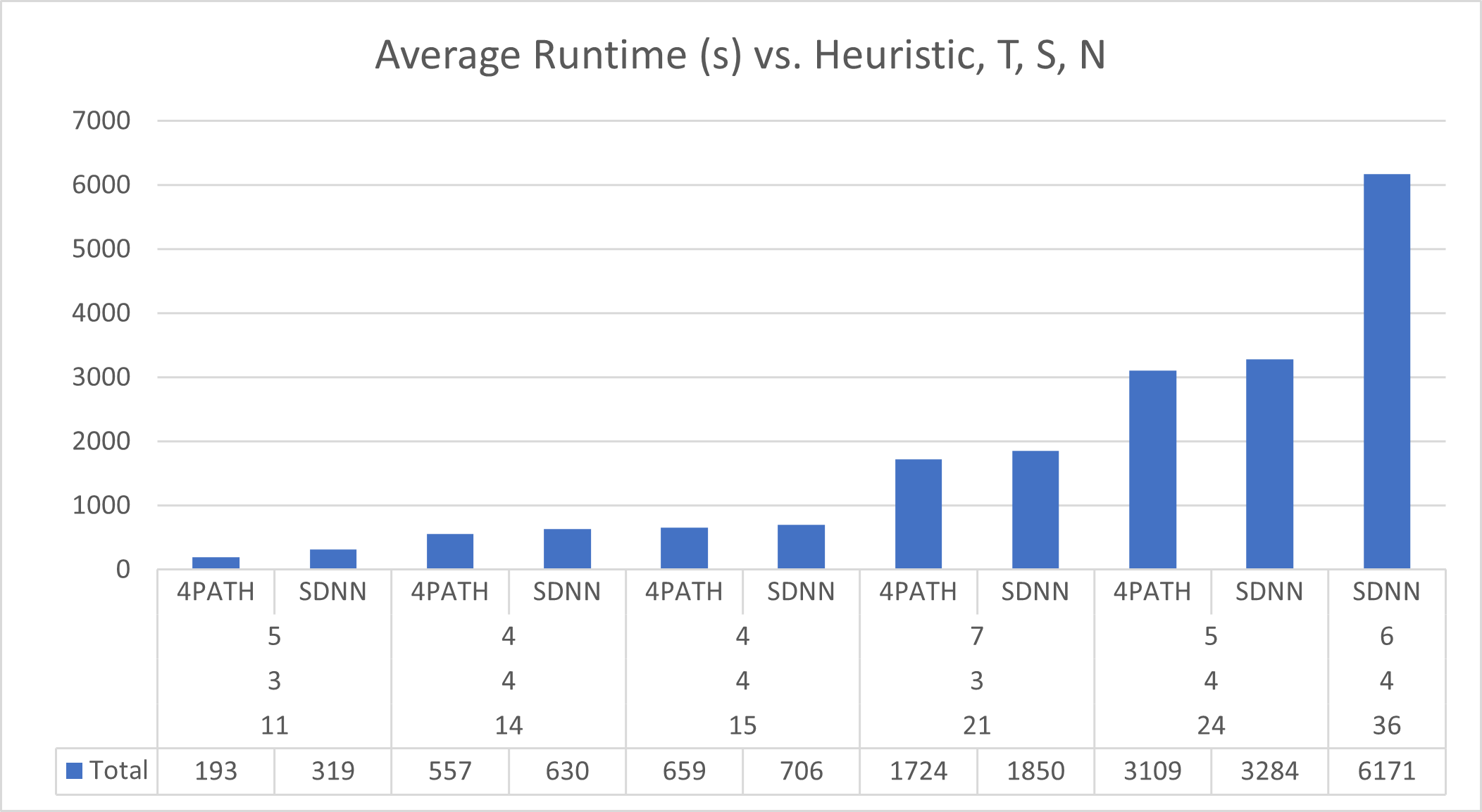}
\caption{Average Runtime For Scaled DNN Heuristic and 4PATH Heuristic}
\label{runtime_v_heur_t_n_s}
\end{figure}

When looking at figure \ref{runtime_v_heur_t_n_s} of runtime for all 1000 iterations for the different N, S, T, and heuristic values, SCALED\_DNN runs more slowly than 4PATH. This difference is marginal when compared to the differences brought upon by different N, S, and T values. Our experimental runtime proves our efficient vectorization technique reduces our practical runtime drastically. $R(3, 7, 21)$ search has a theoretical step runtime of $\binom{21}{7}$, from counterexample checking on every iteration. Likewise $R(4, 4, 15)$ has a theoretical step runtime of $\binom{15}{4}$. Notice $\binom{21}{7}/\binom{15}{4} \approx 85.19$. At 1000 iterations, figure 2 shows that the DNN runtime for $R(3, 7, 21)$ is only $1850/706 \approx 2.62$ times slower than the DNN runtime for $R(4, 4, 15)$, which is $85.19/2.62 \approx 32.52$ times faster than the theoretical comparison. This practical speedup holds for the other values as well.

\subsection{FROM\_CURRENT vs FROM\_PRIOR}

\begin{figure}[ht]
\centering
\includegraphics[width=0.9\columnwidth]{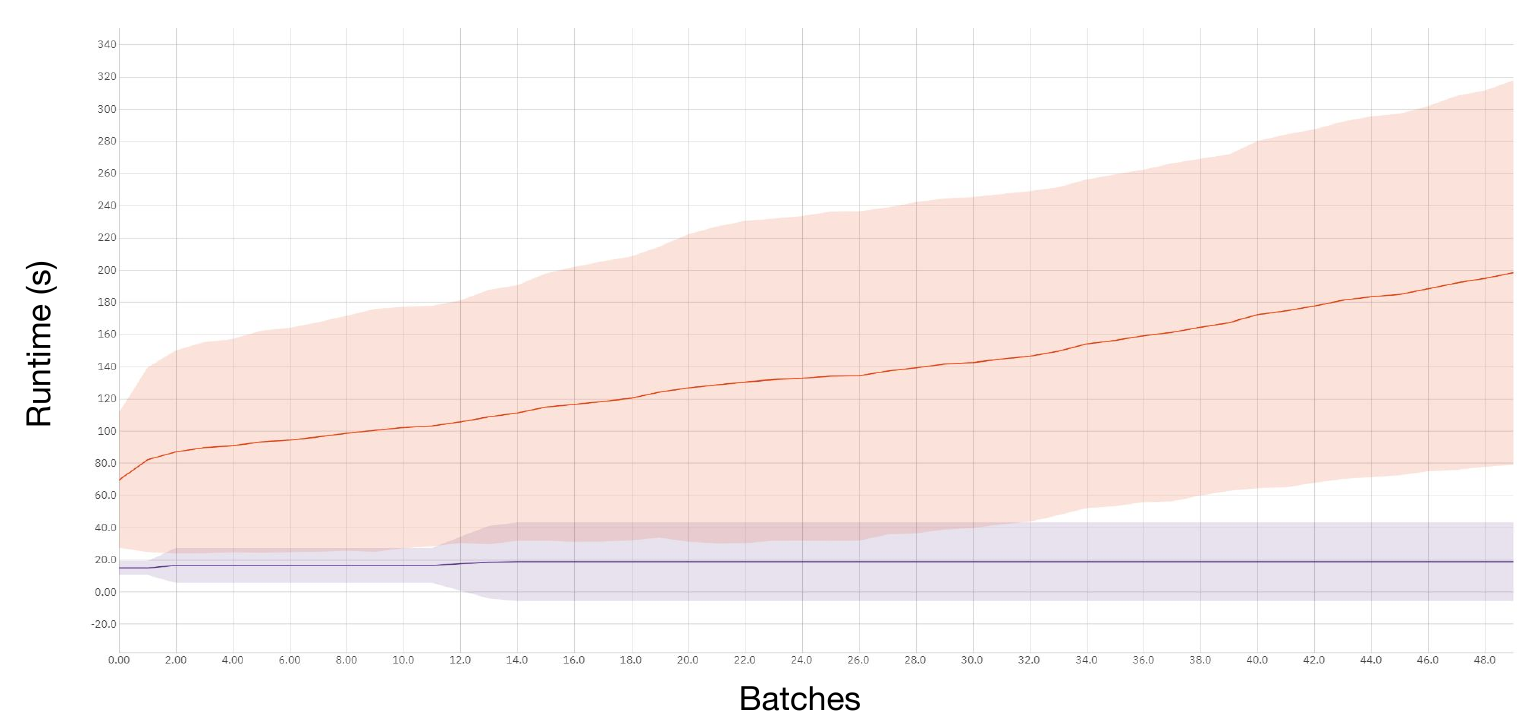}
\caption{Counters found over iterations when grouped by heuristic for $R(4, 4, 14)$ FROM\_CURRENT. The 4PATH heuristic is shown in orange, while the Scaled DNN is in purple.}
\label{running_counter_count_4_4_14}
\end{figure}

\begin{figure}[ht]
\centering
\includegraphics[width=0.9\columnwidth]{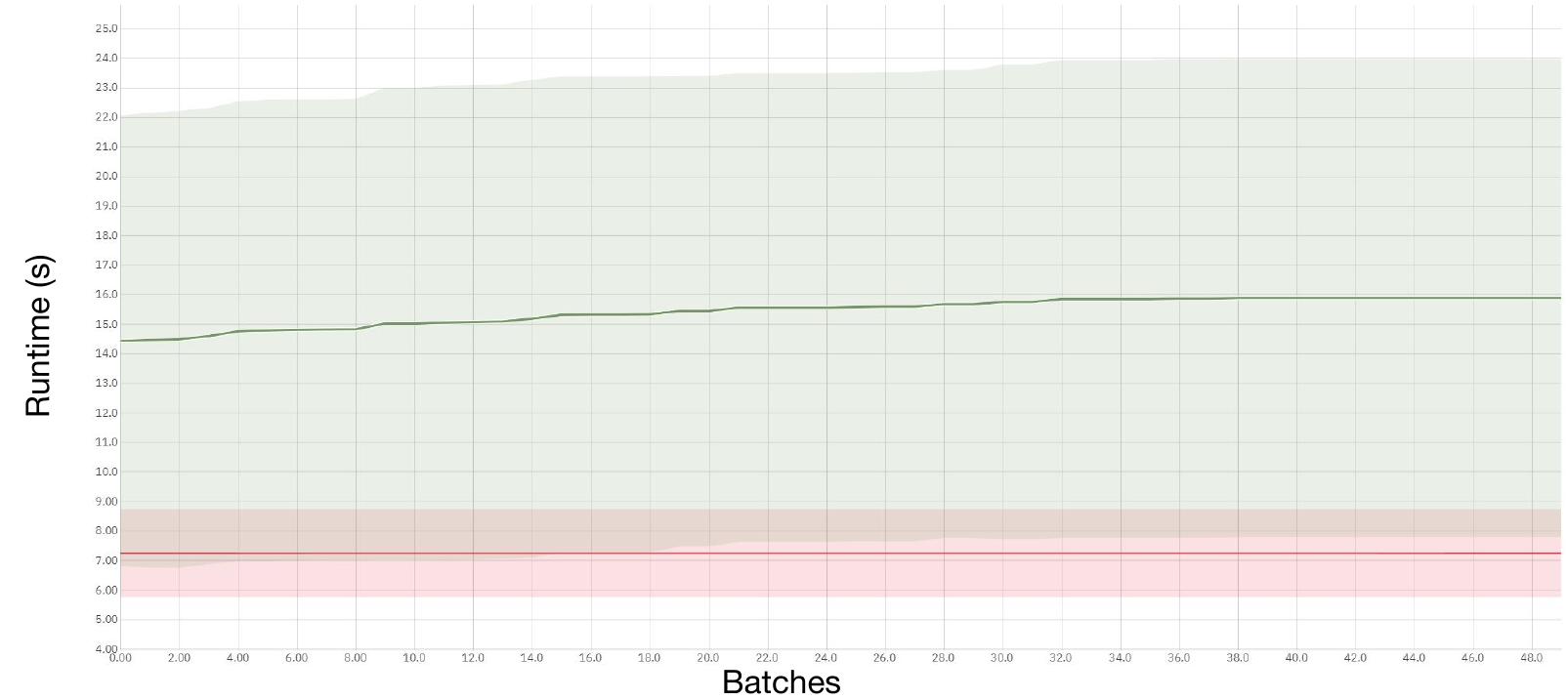}
\caption{Counterexamples found over iterations when grouped by heuristic for $R(4, 4, 15)$ FROM\_PRIOR. The 4PATH heuristic is shown in green, while the Scaled DNN is in red.}
\label{running_counter_count_4_4_15}
\end{figure}

In a general sense, it seems that when using FROM\_CURRENT, the vast majority of counters are found within a single iteration, but a nontrivial amount are found in later iterations. Figure \ref{running_counter_count_4_4_14} reinforces this idea, as do the majority of runs. However, as runs got more complex with increasing N values, the amount of counters found after 100 steps diminished when compared with lesser N values. When using FROM\_PRIOR, the model understandably struggled more greatly with finding counters. Because of how sparse results were, it is difficult to generate any sound figures not affected by the variance of a problem of this sort. Figure \ref{running_counter_count_4_4_15} does show that, nonetheless, counters are found over time. Table \ref{tab:all_runs} also shows that 14 runs were executing for $R(4, 6, 36)$, attempting to push the value. This is not the focus of the paper, though, and so is included for the sake of drawing conclusions on runtime bounds.

Runs are yet to be performed for DNN heuristics other than the simple architecture suggested in the methodology.

Figures \ref{running_time_v_counters}, \ref{running_counter_count_4_4_15}, and \ref{running_counter_count_4_4_14} were generated through Neptune.ai

\section{Conclusion}
\subsection{Bias in Results}
The first thing to clarify in the results is that the runtime for the SDNN should never be faster than 4PATH. This is because the time to compute the heuristic is slower for the DNN than it is to return a single value. Additionally, the model is trained and logged in Neptune.ai every batch of iterations. In the case of our results, this means 50 batches of 20 iterations. 

The motivation of using a reinforcement learning approach is that it will succeed in finding the first nontrivial counterexample more quickly, not that it will iterate more quickly. We did see these results when the model was pretrained on a healthy representative of pretraining data similar to the actual data. As \textit{n} increased, the training data became less and less relevant, and so performance dropped as expected.

Counting the number of counterexamples found, while a productive metric, may also have been misleading. Many of our runs went in and out of dense areas of counterexamples, finding very few or very many over the course of a few steps. This behavior was noted by Exoo for past counterexample values. This behavior should be leveraged by using a more reflective metric when comparing more complex methods.

Regarding complex methods, there are many obvious steps that would lead to improved performance for counterexample searching. Since they minimize how supervised our approach is, they are avoided. Pretraining on all known counterexamples from the highest counterexample value as well as random and nearby negative data points would create a more intelligent searcher. This training set would also see success for a learning rate greater than 0. More complex neural networks may also be beneficial given the compilation of larger training data.

\subsection{Conclusions and Future Work}
In this paper, we share a framework for exploring Ramsey number counterexamples using a best first search algorithm with a runtime polynomial to the number of nodes.

Although not included in methodology, we attempted a multithreaded approach to each iteration of our search algorithm where we parallelized the neighbor vectorization step using python's multiprocessing library. While this approach was slower because of the associated overhead with switching context, in theory, parallelization should speed up our process. Of course, part of this is also the result of the Python GIL, restricting processes to a single thread. These issues, if overcome with a more lightweight language like C/C++ or distributed computing to workaround the GIL, would result in decreased runtime and improved performance. As an example, this project switched from python's networkx graph library to igraph. This led to a 100x speedup using igraph due to its C backend. While past authors used C/C++ for the majority of work done, we still opted to use Python such that further research can be conducted in a more shareable and interpretable manner.

The code for our framework can be found at the following github link: https://github.com/aLehav/RamseyTheoryRL. We share the code with the hope that future researchers can try out new heuristics - either ones based on human observation or entirely new DNN structures - with our best first search approach.
\bibliography{aaai24}

\onecolumn
\section{Appendix}
\subsection{Hyperparameters}
The following table lists the hyperparameters used in our algorithm.

\begin{table*}[h]
    \centering
    \begin{tabular}{llp{10cm}}
        \toprule
        Hyperparameter & Value & Description \\
        \midrule
        Heuristic & RANDOM & Graph heuristic values are Uniform(0,1). \\
        Heuristic & 4PATH & Number of 4 Paths among all 4-node subgraphs\\
        Heuristic & DNN & Neural Network value without n,s,t in vectorization \\
        Heuristic & SCALED\_DNN & Neural Network value with n,s,t in vectorization \\
        Iter\_batch & $x \in \mathbb{Z}$ & Steps to take before updating model data/weights \\
        Iter\_batches & $x \in \mathbb{Z}$ & None if no stopping value, else num. of iter\_batches \\
        Starting\_graph & RANDOM & Random Graph on n nodes\\
        Starting\_graph & FROM\_PRIOR & Counterexample on n-1 nodes with a lonely nth node\\
        Starting\_graph & FROM\_CURRENT & Counterexample on n nodes\\
        Starting\_graph & EMPTY & Empty Graph with n nodes\\
        Starting\_graph\_path & path & Path to starting graph \\
        Starting\_graph\_index & $x \in \mathbb{Z}^{+}$ & Index of starting graph within starting graph index list \\
        Starting\_edges & $x \in \{0, 1\}$ &  If true, Randomly adds edges to the nth node if starting FROM\_PRIOR \\
        Load\_model & $x \in \{0, 1\}$ & If true, loads a past version of the model \\
        Profiler & $x \in \{0, 1\}$ & If true, enables profiler \\
        Pretrain & $x \in \{0, 1\}$ & If true, pretraining occurs \\
        Pretrain\_data & $x$ is a [] & A list of preprocessed csv's to be trained on \\ 
        Training\_epochs & $x \in \mathbb{Z}^{+}$ & If pretraining on data, the number of epochs to train for \\
        Epochs & $x \in \mathbb{Z}^{+}$ & Epochs to update the model for after each iteration of training \\
        Batch\_size & $x \in \mathbb{Z}^{+}$& Batch size for training\\
        Loss\_info & tf.keras.losses & Info regarding loss used, improves logging readability\\
        Alpha & Uniform(0,1) & Learning rate for DNN (i.e. if a counter was found k steps ago, a point that isn't a counter in this iteration would have a value of $\alpha^k$)\\

        Model & tf.model & If passed, when using SCALED\_DNN or DNN heuristic, the default model of a DNN with hidden layers of size 32, 16, and then 1 with optimizer=params['optimizer'], loss=params['loss'], metrics=['accuracy'] can be overridden for any tf model that has already had model.compile called \\
        \bottomrule
    \end{tabular}
    \caption{Hyperparameters used in the experiments, their possible values and descriptions of those values.}
    \label{tab:hyperparameters}
\end{table*}
\end{document}